# Automatic Identification of Document Translations in Large Multilingual Document Collections

**Bruno Pouliquen, Ralf Steinberger & Camelia Ignat**
European Commission - Joint Research Centre
Institute for the Protection and Security of the Citizen (IPSC)
T.P. 267, 21020 Ispra (VA), Italy
http://www.jrc.it/langtech
`Firstname.Lastname@jrc.it`

## Abstract

Texts and their translations are a rich linguistic resource that can be used to train and test statistics-based Machine Translation systems and many other applications. In this paper, we present a working system that can identify translations and other very similar documents among a large number of candidates, by representing the document contents with a vector of thesaurus terms from a multilingual thesaurus, and by then measuring the semantic similarity between the vectors. Tests on different text types have shown that the system can detect translations with over 96% precision in a large search space of 820 documents or more. The system was tuned to ignore language-specific similarities and to give similar documents in a second language the same similarity score as equivalent documents in the same language. The application can also be used to detect cross-lingual document plagiarism.

## 1 Introduction

The task of mining for translational equivalences at document level presented in this paper is based on the automatic mapping of documents onto an existing multilingual knowledge structure, the EUROVOC thesaurus, which is evolving to be a cataloguing standard in parliamentary documentation centres and other institutions in Europe. Before the similarity of documents across languages can be measured, the texts have to be represented by a list of thesaurus descriptor terms. However, EUROVOC descriptor terms are abstract multi-word concepts that can only rarely be found in the documents. In order to identify the best descriptors to represent document contents, we devised an associative system that learns large lists of words that are associated to each descriptor term from a training collection of texts to which descriptors had been assigned manually by professional library indexers. A large overlap of a descriptor's associated words and the words of a text indicates that the descriptor is of some relevance to the text.

After describing the possible uses of a system that can identify translations of a given document (sect. 2), and after relating to previous work in this field (sect. 3), we describe the multilingual knowledge structure EUROVOC used in the process (sect. 4) and the mapping procedure to represent document contents across languages with this resource (sect. 5). In section 6, we present several experiments for cross-lingual document comparison, and the performance achieved. Section 7 summarises the results and points to future work.

## 2 Motivation

Texts and their translations are a rich linguistic resource as they can be used to train and test a number of multilingual and cross-lingual applications such as statistics-based Machine Translation systems, Cross-Lingual Information Retrieval (CLIR) engines, multilingual sentence aligners, named entity recognition and classification tools, and many more. They can furthermore be exploited to generate bilingual or multilingual linguistic resources such as terminology lists, dictionaries, and cross-lingual word associations. For the consistency of these automatically built resources, collections of parallel texts are more useful than several independent or comparable monolingual corpora. For this reason, Resnik (1998) as well as Chen & Nie (2000) have devised systems to automatically compile such parallel corpora from the internet, and they, as well as Smith (2002) and Landauer & Littman (1991) have worked to build applications that try to decide whether texts are translations of each other.

Our own main motivation to carry out this work was to help users find related texts and translations of a given text in large multilingual document collections. The identification of translations and other similar texts can be useful to suggest to users to read texts in a language they may understand better. In a different setting, the inverse may also be true: Users may not want to be presented with documents that are too similar to those that they already know in order to avoid redundant reading. We will also soon need the automatic identification of semantically similar newspaper articles written in various languages to track articles on the same topic (topic tracking) and to detect trends.

Our work can furthermore help to overcome the yet unsolved problem of how to detect document plagiarism across-languages (Clough, 2003). Finally, it should also be possible to use our method to compile *comparable corpora* on the basis of readily available text collections such as those distributed by Reuters and Le Monde. Although this has not yet been tested, it should be possible to take news articles about a certain subject field in one language and to produce a collection of the most similar documents in the other language. These can then be joined to build a comparable corpus of this subject field.

## 3   Related Work

Most previous work on finding translational equivalences concentrates on the sentence or the word level. However, some researchers were interested in compiling automatically large collections of parallel texts to exploit them for applications such as those described in the previous section. Both Resnik (1999) and Chen & Nie (2000) devised systems to mine the internet for parallel texts with the help of conventional search engines by finding parent pages containing HTML anchors such as 'English version' and 'French version' or by searching text in one language containing an anchor pointing to its translation. The result was, for each web site, a small number of HTML documents that are likely to be translations of each other. The challenge was then to determine whether or not these documents were really translations. In order to do this, Resnik applied a language identification tool and used structural (formatting) features to align the two texts, using *diff*. He then used the string length criterion for the aligned chunks, meaning that it was assumed that these chunks were translations of each other if they were of a similar length. Resnik evaluated the results on the basis of 261 document pairs against human judges. He achieved a precision (i.e. the percentage of automatically identified document pairs that were really translations of each other) of 92% with a recall of 64.1%, which means that almost two thirds of the real translation pairs were found.

For a similar task, Chen & Nie used the text features 'URL', 'size', 'date', 'language' and 'character set' to determine whether documents were translations of each other. They evaluated their translation identification algorithm on the basis of 100-200 randomly picked document pairs. Although they did not exploit the expected similar formatting of the text and its translation, which is a major feature in Resnik's work, they report a precision of over 95% for English-French and of 90% for English-Chinese document pairs.

A fundamental difference between the web mining task and ours is that, in the web mining task, the search space for the translation consists of the relatively small number of documents found on the same site, while the search space in our experiments consisted of up to 1615 documents and could be much larger.

Landauer & Littman (1991) used an approach that is more comparable to our own in that it uses lexical text features and that the search space for the translations in their experiment was much bigger. They applied cross-lingual *Latent Semantic Indexing* (LSI) to text segments, i.e. smaller parts of a text. The basic function of cross-lingual LSI is to analyse a training set of parallel texts statistically to derive a common vector space representation scheme for all the words in the bilingual sample, meaning that the words in both languages are represented using the same language-independent vector space. The vector space of typically 100 dimensions is achieved by applying *Singular Value Decomposition* and by going via a term-by-document matrix to several lower rank matrices (see Deerwester et al., 1990). The result of this training phase is that each smaller or larger piece of text in either of the two languages can then be represented using this same vector space. For each pair of documents in these two languages, a cosine similarity score can then be calculated. The higher the similarity, the more similar the texts. Translations are supposedly close to 100% similar. Landauer & Littman's evaluation using the 'Hansard' parallel corpus showed that the average similarity for 200 randomly sampled pairs of English-French paragraphs according to their representation was 0.78 (with a standard deviation of 0.09). Their evaluation on a collection of 1,582 parallel text paragraphs showed that their LSI system managed to find the French translation as the most similar text to a given English paragraph in 92% of all cases (92% precision), which is a very good result.

Although the LSI approach is rather different from our thesaurus-based one, the results achieved are very similar to the ones presented in this paper. The advantage of Landauer & Littman's approach is that their system can theoretically be trained for any language pair or set of languages, while our own is limited to the languages for which EUROVOC and pre-indexed training material exists. The advantage of our own approach is that it is more modular and that no language-pair specific training is required because the multilingual conceptual knowledge structure EUROVOC provides the link between the languages once the descriptors have been assigned to texts in a language. Landauer & Littman's approach needs retraining each time a language is added or the training set changes. Another advantage of our approach is that the representation of texts by their most appropriate EUROVOC descriptors fulfils many more uses than the identification of translations. These include subject-specific summarisation, conceptual and cross-lingual indexing for human users, and automatic annotation of documents for the Semantic Web (Pouliquen et al., 2003).

| Lemma | Weight |
|---|---|
| dangerous_goods | 33 |
| radioactive_material | 19 |
| by_road | 19 |
| carriage | 19 |
| dangerous | 18 |
| plutonium | 17 |
| radioactive_waste | 15 |
| nuclear_fuel | 15 |
| shipment | 15 |
| adr | 14 |
| bind_for | 13 |
| tank | 13 |
| receptacle | 13 |
| transport | 13 |
| pollute | 12 |
| nuclear_waste | 12 |

**Table 1**. Most important English associated lemmas for EUROVOC descriptor TRANSPORT OF DANGEROUS GOODS.

Clough (2003) and Shivakumar & García-Molina (1996) focused on monolingual document similarity calculation in order to detect plagiarism. While Shivakumar & García-Molina mainly applied a cosine similarity measure to various chunks of two documents, Clough applied sort of a cosine similarity to a variety of word n-grams and looked at rare words and stylistic inconsistency between paragraphs. Plough points out the difficulty of identifying more complex forms of plagiarism that go beyond cut-and-paste or simple rewriting. Furthermore he stresses the difficulty of detecting cross-lingual plagiarism (Multilingual Copy Detection). We believe that our approach provides at least partial solutions for the multilingual plagiarism of whole documents or of document sections (after splitting documents into smaller parts).

## 4 EUROVOC Thesaurus

EUROVOC is a hierarchically organised controlled vocabulary that was developed by the European Parliament, the European Commission's Publications Office, together with many national parliaments in and outside the European Union (EU), for the cataloguing, search and retrieval of their large multilingual document collections (Eurovoc 1995). Currently, the cataloguing is a manual (intellectual) process. EUROVOC is a steadily growing resource that is currently available in the eleven official EU language versions with an additional eleven languages in preparation.

Its 6075 descriptor terms are hierarchically organised into 21 fields and, at the second level, into 127 micro-thesauri. The maximum depth is eight levels. In addition to the 5877 pairs of broader terms (BT) and narrower terms (NT), there are 2730 pairs of related terms (RT) linking descriptors not related hierarchically.

Due to its wide coverage represented by the relatively small number of descriptors, the descriptor terms are mostly rather abstract multi-word expressions that are unlikely to be found verbatim in the texts. EUROVOC examples are PROTECTION OF MINORITIES, FISHERY MANAGEMENT and CONSTRUCTION AND TOWN PLANNING[1].

## 5 Mapping Documents to EUROVOC

The procedure of mapping texts written in different languages automatically onto the multilingual EUROVOC thesaurus is described in detail in Pouliquen et al. (2003) so that this process will only be sketched here. A description of previous results and experiments can be found in Steinberger (2001).

It is not possible to base an automatic EUROVOC thesaurus descriptor assignment on the actual occurrence of the descriptor text in the document because the lexical evidence is weak and even misleading: The descriptor text occurs explicitly in only 31% of documents manually indexed with this descriptor. On the other hand, in approximately nine out of ten documents, the descriptor text is present explicitly even though the descriptor itself had *not* been assigned manually. This means that basing the assignment on the presence of the descriptor text in the document would lead to wrong results in most cases.

For this reason, we have taken another, fuzzy approach. For each descriptor, we automatically produce its profile, i.e. a ranked set of words that are statistically related to the descriptor so that we have more lexical evidence at hand that indicates that a certain descriptor should be assigned to a text. As these words are statistically, but not always semantically related, we refer to these pertinent words as descriptor *associates*. Table 1 shows the top of the relevance-ranked associate list for the descriptor TRANSPORT OF DANGEROUS GOODS.

---
[1]We write all EUROVOC descriptors in small caps.

We produce these associate lists on the basis of a large collection of manually indexed documents (the *training set*), by comparing the word frequencies in the subset of texts that have been indexed manually with a certain descriptor with the word frequencies in the whole training set. For the comparison, we use a combination of Dunning's statistical log-likelihood test (or $G^2$) to reduce the number of words to be considered (dimensionality reduction) with filters and various IDF (Inverse Document Frequency) weightings (Salton & Buckley 1988). This method automatically produces lists of typical words for each descriptor. It also produces information on the degree with which these words are typical.

Before applying any statistical procedures, we carry out some minimal linguistic pre-processing in order to normalise all texts: We lemmatise texts so as to work only with base word forms, and we mark up the most frequent compounds in our training set. There is also a large list of stop words, i.e. words that should never be associates because they are not meaningful.

When we want to index a new text automatically with EUROVOC descriptors, we make a statistical comparison of the frequency list of its lemmas with the associate lists of all descriptors to check which associate lists are most similar to the text's lemma list. The most similar associate lists, according to some statistical similarity measure, indicate the most appropriate descriptor terms. The result is thus a long ranked list of EUROVOC descriptors assigned to this document. Typically, we keep the highest-ranking 100 descriptors. The example in Table 2 shows the assignment results of a document found on the internet[2].

To speak with text classification terminology (Sebastiani, 1999), the mapping of texts onto the EUROVOC thesaurus is a category ranking classification task using a machine learning approach, where an inductive process builds a profile-based classifier by observing the manual classification on a training set of documents with only positive examples. For each EUROVOC descriptor, we built a category profile consisting of a set of lemmas and their weight. Unlike usual classifiers, our system does not decide against the appropriateness of a class. Instead, it produces long ranked lists of more or less relevant classes for each document. This representation is more suitable for document similarity calculation. According to Sebastiani (1999), the k-nearest-neighbour (KNN) approach tends to produce best results. Although we have not tried this approach, it does not seem computationally viable to apply it to our training set of over thirty thousand documents, and we doubt that it would be the most appropriate technique for our multi-class categorisation problem.

---
[2] http://europa.eu.int/comm/food/fs/inspections/vi/reports/austria/vi_rep_oste_1074-1999_en.html

| Rank | Descriptor | Cosine |
|---|---|---|
| 1 | VETERINARY LEGISLATION | 42.4% |
| 2 | PUBLIC HEALTH | 37.1% |
| 3 | VETERINARY INSPECTION | 36.6% |
| 4 | FOOD CONTROL | 35.6% |
| 5 | FOOD INSPECTION | 34.8% |
| 6 | AUSTRIA | 29.5% |
| 7 | VETERINARY PRODUCT | 28.9% |
| 8 | COMMUNITY CONTROL | 28.4% |

**Table 2**. Assignment results (top 8 descriptors) for a document found on the internet ('Food and veterinary Office mission to Austria')[2]

The method used is language-independent. It has been applied to all eleven official EU languages with similar results, but the parameter settings have currently only been optimised for English and Spanish, and to a lesser degree for French. Further languages will follow. As EUROVOC is used by parliaments in several Eastern-European countries that will join the EU soon, we are currently trying to get collections of manually indexed documents to train our system for these languages.

The system was trained on 30231 texts and it was evaluated on a complementary test set of 590 representative, but randomly chosen texts. The descriptor assignment was evaluated against the annotation of two separate indexing professionals. The precision achieved for the eight highest-ranking descriptors assigned automatically (eight was the average number of descriptors assigned manually) was 67% (F=65). This is 86% as good as the inter-annotator agreement of 78% (67/78=86%).

The assignment of descriptors to a text of one to five pages currently takes ten to twenty seconds on an average modern PC (1.3 GHz AMD processor with 384MB of RAM), but the process could be optimised.

## 6 Identification of Translations

Each EUROVOC descriptor is identified by a numerical code and has one-to-one translations into different languages. Texts can be represented as a vector consisting of the EUROVOC descriptor codes assigned to them and their assignment score. Two texts can then be compared with each other by calculating the similarity between the two descriptor vectors. The higher the similarity value, the more similar the texts. The similarity measure for documents is the same, independent of the document language.

### 6.1 Description of the Test Sets

We tested the algorithm on three different test sets of between 795 and 1615 document pairs each. Neither of the test sets was part of the training corpus or of the set of documents used to optimise the EUROVOC assignment algorithm.

Test set T0 consists of 1130 randomly chosen texts of the same type as the training data. As T0 contained many almost identical documents (e.g. updates to the same contract in which only the dates have changed, or the same international agreement of the EU with two different countries in which the largest parts of the texts were identical copies), we identified documents that were 95% or more identical in English and took them and their translations out of the corpus. This means that we took those documents away that were so similar that most parts of the text were verbatim the same. The resulting test set T1 contained 820 documents. Test set T2 consists of 795 official EU bulletins[3], a text type our system had not previously been exposed to. To test corpus-specific behaviour, we also merged T1 and T2 into T3, resulting in 1615 document pairs.

### 6.2 Semantic Text Similarity

We represent each text as a vector of the top 100 automatically assigned EUROVOC descriptors and their assignment score, and we calculate the cosine (Salton & Buckley, 1988) between vectors to measure text similarity. The number of dimensions of the vector used in the similarity calculation is the union of the sets of dimensions of the two individual vectors. In the case of two entirely unrelated documents (no descriptors in common, similarity equals zero), the number of dimensions would thus be 200.

In the first experiment, we checked for each English document whether the Spanish and French translations were found as the most similar documents by looking only at the texts in the target language. This means that, when searching for the Spanish translations of the 820 English documents of T1, we looked only at the 820 Spanish documents. Table 3 shows for T1 that 90.61% of all Spanish translations were found as the most similar document to the English one (row T1-ES). Searching in the other direction, i.e. searching English translations of Spanish texts (T1-SE), produces results that are about 2% lower. We have not yet been able to identify the reason for this difference in performance. The results for French (T1-EF) are slightly lower because the French EUROVOC descriptor assignment has not been optimised entirely.

The comparison of T1-ES with T2-ES, and of T1-EF with T2-EF, shows that the system performs better with test set T1, i.e. if the test texts are of the same type as the training set. However, the precision of 84,28% for test set 2 (T2-ES) is still high.

The results for test set T0 are displayed for completeness (T0-ES) to show the decrease in performance when the many monolingual document duplicates have not been removed.

### 6.3 Considering the Length Factor

These results were produced exclusively on the basis of semantic similarity, i.e. on the basis of the EUROVOC content descriptors, not taking into account the obvious fact that translations should have a similar length to the original text. In our training set, we observed that the Spanish and French translations use, on average, 13.5% and 18% more characters, respectively, than their English counterparts. The variation of the length difference approximately follows a normal distribution. In the second experiment, we considered this length factor LF by punishing the similarity score $\cos(d_1,d_2)$ of all those documents $d_1$ and $d_2$ that differ from this average length difference according to the following formula (with μ and σ being the average and the standard deviation of the length factor on the training set):

$$Simlf(d_1,d_2) = e^{-0.5\left(\frac{\frac{length_{d1}}{length_{d2}}-\mu}{\sigma}\right)^2} \cos(d_1,d_2)$$

Row T1-ES shows that adding this criterion helps to increase the translation identification precision rate from 90.61 to 96.83%. Most of the remaining translations were identified as the second or third most similar document: 99.76% of all translations (all but two) were found as one of the three most similar documents; the remaining two were found at ranks 7 and 16. We consider these to be very good results for the large search space of 820 documents. Applying the length factor alone, i.e. without EUROVOC descriptor assignment, raises the precision for random selection of the translation from 0.12% to 1.7% (T1-ES-LF), which is a very bad result and which confirms that the Length Factor alone cannot be used to detect translations.

Friburger & Maurel (2002) showed that cross-lingual information retrieval can be improved by considering named entities. We expect that our translation identification results can be improved further by considering named entities and additional text features such as formatting, the length of individual text passages and the occurrence of numbers, dates and cognates in the text. However, we restricted ourselves to using only the 'semantic' similarity to show the benefit of this approach. Furthermore, using cognates is character set-dependent and would not work for English-Greek while the EUROVOC assignment approach is not restricted in this way.

---

[3] Found at http://europa.eu.int/abc/doc/off/bull/en/welcome.htm

| Experiment | | Translations found as most similar document | |
|---|---|---|---|
| Identifier | Explanation | No length factor | Length factor |
| T0-ES | Finding translations of 1030 English texts in a collection of 1030 Spanish texts | 81.45% | 90.26% |
| T1-ES | Finding translations of 820 English texts in a collection of 820 Spanish texts | 90.61% | 96.83% |
| T1-SE (Es to En) | Finding translations of 820 Spanish texts in a collection of 820 English texts | 88.05% | 94.76% |
| T1-EF | Finding translations of 820 English texts in a collection of 820 French texts | 81.71% | 91.46% |
| T2-ES | Finding translations of 795 English texts in a collection of 795 Spanish texts (EU bulletins) | 84.28% | 90.31% |
| T2-EF | Finding translations of 795 English texts in a collection of 795 French texts (EU bulletins) | 80.00% | 84.91% |
| T1-ES-LF | Finding translations of 820 English texts in a collection of 820 Spanish texts using only the length factor (no semantic similarity) | 00.12% | 01.71% |
| T3-ES (T1+T2) | Finding translations of 1615 English texts (820 T1 + 795 T2) in a collection of 1615 Spanish texts (combining 'T1' and 'T2' sets) | 85.08% | 93.44% |
| T1-ES-B (bilingual) | Finding translations of 820 English texts in a collection of 1640 Spanish and English texts | 62.56% | 77.32% |
| Th1-ES-B (weighted-bilingual-half) | Finding translations of 410 English texts in a collection of 820 Spanish and English texts (half the set of 'T1-ES-B', random-selected) | 69.68% | 81.91% |
| T1-ES-BW (bilingual-weighted) | Finding translations of 820 English texts in a collection of 1640 Spanish and English texts. Punishing the mono-lingual similarity score by 83% | 89.27% | 95.24% |
| Th1-ES-BW | Finding translation of 410 English texts in a collection of 820 Spanish and English texts. Punishing the mono-lingual similarity score by 83% | 92.91% | 96.82% |

**Table 3**. Translation recognition results (Precision), i.e. number of translations that were identified as being the most similar text in the test set, for various test sets (T0, T1, T2, T3), languages (English E, Spanish S and French F), language directions (ES vs. SE), with and without applying the length factor LF.

The success rate of 90.31% for test set T2 was lower (row T2-ES), which shows that the system works better on the sort of texts it was trained on. We experienced that EUROVOC descriptor assignment seems less good when applied to news articles or other documents that are completely different from the training documents, and especially if they cover subjects not well represented by EUROVOC (e.g. trivia, sports, etc.). We therefore expect that document similarity calculation for such documents would produce less good results, as well.

We also tested the performance on the combined set of 1615 T1 and T2 texts, i.e. searching for the Spanish translation of English texts of both test sets in both test sets (T3-ES). The success rate of 93.44% is surprisingly high as it seems to be simply an average of the performance of the T1 (96.83%) and the T2 sets (90.31%). We expected a lower precision because the search space is twice as big as in T1 and T2. A possible explanation for this is that the two collections overall have different semantic contents so that the risk of wrongly identifying a translation from the other text type is rather low. It is also possible that the specific idiosyncratic features of each of the text types have an impact on the assignment so that texts of the same type are more likely to get similar EUROVOC descriptors assigned than other texts with similar content.

### 6.4 Cross-lingual vs. Monolingual Similarity

As the training and assignment of EUROVOC descriptors to text is done individually and separately for each language, the assignment results differ from one language to the other. While the (monolingual) similarity of a text to itself is 1.0, the similarity of all English-Spanish translation pairs in a set of 590 document pairs not being part of the training or the test set is 0.83, with a standard deviation of 0.09.[4] It is therefore possible that our system will identify similar documents written in the same language as more similar than the translation in another language, while we would like our system to evaluate content similarity without being biased by the document language.

---
[4] This compares to Landauer & Littman's (1991) average translation pair similarity of 0.78, as described in sect. 3. Ideally, translation pairs should be identified as 100% similar.

To test this, we launched the similarity measure (including the Length Factor) on the bilingual English-Spanish test set T1 (i.e. 820 English plus 820 Spanish texts). For each English text, the most similar document was identified within the whole search space of 1640 English and Spanish documents. Row T1-ES-B (bilingual) in Table 3 shows that the performance is indeed much lower (77.32% vs. 96.83% monolingually), but the results are difficult to compare because the bilingual search space is twice as big as the monolingual search space. To make the results comparable, we launched the same experiment for a random selection of 50% of the T1 texts (Th1). However, the results shown in row Th1-ES-B still only go up to 81.91% and are thus about 15% lower than when looking only at the Spanish search space. This indicates that the system is indeed biased towards documents of the same language.

To correct this monolingual bias, we punished the English documents by multiplying the similarity score of English documents with a factor of 0.83 (which is the average similarity of English texts and their Spanish translations in our training set). The idea is that, after this punishment, the English document itself and its Spanish translation should, on average, have the same score. The result of 95.24% precision in row T1-ES-BW (bilingual, weighted) shows that this monolingual bias can indeed be corrected because the result is comparable with the 96.83% achieved when searching only the 820 Spanish documents. When using the same reduced search space of 820 documents (row Th1-ES-BW), the system even achieves 96.82% precision, which is almost identical to T1-ES.

The formula to punish the similarity of the English documents works rather well, but it still needs to be improved. The reason for this is that, currently, a Spanish document with a similarity score higher than 0.83 (e.g. 0.9) will end up having a higher similarity score than the English document itself, which will have a similarity score of 0.83. Instead of the current formula, we plan to apply a more sophisticated function that smoothes the resulting similarity scores.

### 6.5 Is there a translation?

In the previous experiments, each source document had a translation in the search space and the task was to identify it among all documents. However, in a different setting, where it is not known whether there is a translation or not, a similarity threshold has to be set to decide whether or not there is a translation: Only if a document in the target language is more similar than the threshold, we can assume that it is a translation of the source document.

We experimented with various thresholds and found out that the threshold is dependent on the test set. The more similar the documents are to the training set, the higher the average similarity of translation pairs. The average similarity of translation pairs in T1, which is similar to the training set, is 0.82, while in T2 it is 0.79.

The challenge of the task is to decide on a threshold that is low enough to include most translations while being high enough to exclude other documents that are not translations. Looking at T1, if we aim at including 90% of the translations, the similarity threshold has to be set to 0.70. With this threshold, our system recognises 88.8% of all translations successfully, and 2.2% of non-translations are wrongly identified as translations because they passed the threshold and ranked highest (noise). In the different test set T2, however, the same similarity threshold of 0.70 includes only 76.5% of translations (i.e. 23.5% of all real translations would not even be considered as translations). The error rate of documents wrongly identified as translations is 5.0%.

When searching in T2 for translations of the documents from T1, i.e. when searching for translations where there are not any, the threshold of 0.70 produces 4.15% of noise, i.e. of documents wrongly identified as translations.

To summarise: Depending on whether it is more important to increase the recall for correctly identified translations or to decrease the noise of wrongly identified translations, the threshold should be set lower or higher. In any case, it is advised to review the threshold depending on the document collection: It can be set higher for document collections more similar to the training set and should be set lower when dealing with texts that are of a different type.

## 7 Conclusion and Future Work

We have shown that the automatic mapping of documents in different languages to the same multilingual thesaurus EUROVOC produces a language-independent document representation that lends itself very well to monolingual and cross-lingual document similarity calculation. Assuming that the translation of a text should be the most similar text in a document collection, we carried out tests to search for the Spanish and French translations of English documents, using several parallel text corpora ranging from 795 text pairs to 1130 text pairs and searching in a search space of up to 1640 documents. The ex-periments showed that in over 90% of cases, the Spanish translation was found to be the most similar text to a given English text, in a search space of 820 candidates. When adding a Length Factor to punish those documents that diverge from the expected document length, results of almost 97% precision were achieved, meaning that in 97% of cases, the system identified the correct Spanish translation among 820 texts. When comparing the cross-lingual document similarity capacity with the monolingual capacity by searching for the most similar document in the search space of the combined Spanish and English documents, the success rate dropped to 82%, indicating that the system sometimes wrongly identified similar English documents as being more similar than the Spanish translation. By correcting this bias with a monolingual punishment factor, it was possible to raise the translation identification precision to 97%, meaning that the final similarity calculation was shown to be as good as language-independent. These results are extremely encouraging, but also surprising because the EUROVOC assignment performance, on which the similarity measure is based, is lower.

Future work will include applying the algorithm to texts in further languages and, more importantly, to different collections of parallel texts. We are particularly interested in testing the application for monolingual and cross-lingual plagiarism detection. Regarding monolingual plagiarism detection, we expect our system to work better than string or n-gram comparison in cases where the plagiarist has put effort into paraphrasing the text so as to avoid being detected. An experiment with cross-lingual plagiarism would be attractive because currently no solution to this task exists (Clough, 2003).

Should the need arise to improve the task of translation identification further, we will add other text features to the text comparison algorithm. These include document formating, the occurrence of cognates and of automatically extracted references to named entities such as dates and currency expressions, as well as geographical places and other proper names extracted automatically (Ignat et al., 2003).

## Acknowledgements

We would like to thank the European Parliament and the European Commission's Office for Official Publications OPOCE for providing us with EUROVOC and the training material. We are also grateful to the three anonymous reviewers for their useful comments.

## References

Chen J. & J-Y Nie (2000). *Parallel Web Text Mining for Cross-Language IR*. Algorithmica, vol. 28-2: 217-241.

Clough P. (2003). *Old and new challenges in automatic plagiarism detection*. National Plagiarism Advisory Service; http://ir.shef.ac.uk/cloughie/index.html

Deerwester S., S. Dumais, G. Furnas, T. Landauer & R. Harshman (1990). *Indexing by Latent Semantic Analysis*. Journal of the American Society for Information Sciences, 41: 391-407.

EUROVOC (1995). *Thesaurus Eurovoc - Volume 2: Subject-Oriented Version. Ed. 3/English Language*. Annex to the index of the Official Journal of the EC. Luxembourg. http://europa.eu.int/celex/eurovoc

Friburger N. & D. Maurel (2002). *Textual Similarity Based on Proper Names*. Proc. of the workshop *Mathematical/Formal Methods in Information Retrieval* (MFIR'2002) at the 25th ACM SIGIR Conference, pp. 155-167. Tampere, Finland.

Ignat Camelia, Bruno Pouliquen, António Ribeiro & Ralf Steinberger (2003). *Extending an Information Extraction Tool Set to Central and Eastern European Languages*. Proceedings of the RANLP'2003 workshop 'Information Extraction for Slavonic and other Central and Eastern European Languages'. Borovets, Bulgaria.

Landauer Thomas & Michael Littman (1991). *A Statistical Method for Language-Independent Representation of the Topical Content of Text Segments*. Proceedings of the 11th Int. Conf. Expert Systems and Their Applications, vol. 8: 77-85. Avignon, France.

Pouliquen Bruno, Ralf Steinberger & Camelia Ignat (2003). *Automatic Annotation of Multilingual Text Collections with a Conceptual Thesaurus*. Proceedings of the EUROLAN workshop 'Ontologies and Information Extraction'. Bucharest, Romania.

Resnik Philip (1999). *Mining the Web for Bilingual Text*. Proceedings of ACL. Maryland.

Salton G. & C. Buckley (1988). *Term-weighting approaches in automatic text retrieval*. Information Processing & Management, 24 (5): 513-523.

Sebastiani Fabrizio (1999). *A Tutorial on Automated Text Categorisation*. In Analia Amandi and Alejandro Zunino (eds.): Proceedings of ASAI-99, 1st Argentinean Symposium on Artificial Intelligence, Buenos Aires, Argentina, pp. 7-35.

Shivakumar N. & H. García-Molina (1996). *Building a Scalable and Accurate Copy Detection Mechanism*. Proceedings of the 1st ACM Conference on Digital Libraries (DL'96). Bethesda, Maryland.

Smith Noah (2002). *From Words to Corpora: Recognizing Translation*. Proceedings of EMNLP'2002. Philadelphia, Pennsylvania.

Steinberger Ralf (2001). *Cross-lingual Keyword Assignment*. Proc. of SEPLN'2001. Procesamiento del Lenguaje Natural, Revista No 27: 273-280. Jaén, Spain.